\documentclass{INTERSPEECH2023}

\usepackage{amsmath}
\usepackage{amssymb}
\usepackage{mathtools}
\usepackage{amsthm}

\usepackage{tabularx,colortbl}
\usepackage{multirow}
\usepackage{multicol}
\usepackage{textcomp}
\usepackage{hyperref}
\hypersetup{
    colorlinks=true,
    linkcolor=blue,
    filecolor=magenta,      
    urlcolor=cyan,
    pdftitle={Overleaf Example},
    pdfpagemode=FullScreen,
    }
    
\interspeechcameraready

\title{MT-SLVR: Multi-Task Self-Supervised Learning for Transformation In(Variant) Representations}

\name{Calum Heggan$^1$, Tim Hospedales $^2$, Sam Budgett$^3$, Mehrdad Yaghoobi$^1$}

\address{
  $^1$ School Of Engineering, University of Edinburgh, Scotland,\\
  $^2$ School Of Informatics, University of Edinburgh, Scotland,\\
  $^2$ Thales UK RTI}
\email{s1529508@ed.ac.uk, t.hospedales@ed.ac.uk, Samuel.BUDGETT@uk.thalesgroup.com, m.yaghoobi-vaighan@ed.ac.uk}

\begin{document}

\maketitle
\begin{abstract}
Contrastive self-supervised learning has gained attention for its ability to create high-quality representations from large unlabelled data sets. A key reason that these powerful features enable data-efficient learning of downstream tasks is that they provide  augmentation invariance,  which is often a useful inductive bias. However, the amount and type of invariances preferred is not known apriori, and varies across different downstream tasks. We therefore propose a multi-task self-supervised framework (MT-SLVR) that learns both variant and invariant features in a parameter-efficient manner. Our multi-task representation provides a strong and flexible feature that benefits diverse downstream tasks. We evaluate our approach on few-shot classification tasks drawn from a variety of audio domains and demonstrate improved classification performance on all of them.
\end{abstract}


\section{Introduction}
Few-shot learning, which aims to learn with limited data, has become increasingly popular in response to the lack of large labelled datasets for many practical applications. Models trained using self-supervision (where a deep neural network (DNN) is trained with pseudo-labels that define pre-text tasks) have demonstrated strong success on few-shot learning tasks, with contrastive objectives among the most successful. Contrastive methods' efficacy is attributed to learning an inductive bias in the form of invariances to applied augmentations \cite{simclr,linus_invariance}. For example, affine transformation invariance is typically useful for object category recognition, where pose is a nuisance factor \cite{simclr}. However, the ideal type and degree of invariance is not known apriori, and varies across downstream tasks. So, contrastively trained invariant features do not provide a \emph{one size fits all} solution \cite{linus_invariance,aug_self,hyper_simclr}. For example, a model learned to be pitch-shift invariant \cite{clar} would likely fail a task which relies on pitch sensitivity features. To learn a model which can successfully solve various downstream tasks, we require a feature representation with both invariant and transformation-sensitive properties.

\noindent We propose a parameter-efficient multi-task learning framework to address this limitation of existing contrastive learners. We simultaneously learn a contrastive objective (to learn augmentation invariances) and a transformation prediction objective (to learn augmentation sensitivity), thus providing a more flexible feature for downstream tasks. 
Our contributions include: 1) A novel multi-task learning  framework; 2) A parameter-efficient solution to multi-task learning based on task-agnostic and task-specific features; 3) Evaluation of few-shot classification over 10 datasets, spanning audio and speech domains; 4) Analysis of learnt invariance strength and its relation to performance.  \href{https://github.com/CHeggan/MT-SLVR}{Code can be found here}.

\section{Self-Supervision for Few-Shot Classification}
A common goal of using self-supervision is to learn a powerful data representation without the need for large corpuses of labelled training data. This representation can then be used for other downstream tasks, where it can either be fine-tuned, using some labelled data from the target domain, or left as a static feature extractor. This type of approach is a particularly strong candidate for use in few-shot learning, where training a model from scratch for the task is difficult due to limited amount of labelled examples. 

\noindent This use case of self-supervision is utilised in our work. In particular, we use pre-trained self-supervised models (used as static feature extractors) and a linear classifier in order to solve few-shot classification tasks. 

\noindent Such few-shot problems can be formalised in terms of containing a support set $\mathcal{S}$ with a few training samples per class and a query set $\mathcal{Q}$ with test samples. These tasks are typically expressed as N-Way K-Shot tasks, with N being the number of classes and K being the number of examples per class. More formally, task components look like:
\begin{equation}
    \mathcal{S}=\left\{\left(x_{1}, y_{1}\right),\left(x_{2}, y_{2}\right), \ldots,\left(x_{\mathcal{M}}, y_{\mathcal{M}}\right)\right\}
    \label{eq:support_set}
\end{equation}
\begin{equation}
    \mathcal{Q}=\left\{\left(x_{1}, y_{1}\right),\left(x_{2}, y_{2}\right), \ldots,\left(x_{\mathcal{L}}, y_{\mathcal{L}}\right)\right\}
    \label{eq:query_set}
\end{equation}
where each example $(x, y)$ consists of an input $\mathbf{x} \in \mathbb{R}^{D}$ and a class label $\mathbf{y} \in \{1, \ldots, N\}$, with $\mathcal{M}$ and $\mathcal{L}$ being the total number of support and query examples respectively.


\section{Related work}
\textbf{Self-Supervised Learning:} Since self-supervised learning is a large topic \cite{linus_survey,ssl_audio_survey,ssl_speech_survey}, we focus on relevant trends for brevity. One key trend is the success of methods which utilise augmentations for learning, including many contrastive methods \cite{simclr,simsiam,moco,barlow_twins} as well as predictive ones \cite{rot_net}. These approaches learn invariances or sensitivity to applied augmentations, respectively. Audio-specialised contrastive variants include COLA \cite{cola}, CLAR \cite{clar}, and the work by Fonseca et al. \cite{ucl_ser}. In this work, we focus on SimCLR \cite{simclr}, SimSiam \cite{simsiam} and a custom transformation prediction framework. In the SimCLR/SimSiam methods, augmentation pipelines generate multiple 'views' of each data point and the DNN is trained to map them to a similar area in the feature space, allowing the model to learn an augmentation-invariant representation. SimCLR \cite{simclr} and SimSiam \cite{simsiam} are distinct in a few ways; SimCLR uses implicit negative sampling and a temperate scaled cross-entropy loss, while SimSiam only uses positive-pair contributions and optimizes for cosine similarity. Utilising the same augmentation pipelines as described above, transformation prediction algorithms instead try to predict how or if specific augmentations have been applied to input samples \cite{rot_net}. Unlike contrastive learning, algorithms with this objective learn sensitivity to augmentations. Since multiple augmentations can be applied to each input, we implement a multi-label TP model, where each augmentation is predicted independently.

\noindent\textbf{Few-Shot Classification for Audio \& Speech:} Currently, only a handful of works exist investigating the few-shot learning regime for acoustic data \cite{transient, fs_fsd, metaaudio}. Within these, few-shot speech classification, especially over different types of speech (language, accent, emotion etc) is heavily underrepresented. We make extensive use of the MetaAudio \cite{metaaudio} benchmark due to its publicly available codebase. Additionally, we propose an extension to MetaAudio, including 3 new speech datasets suitable for few-shot classification \cite{cremad, saa, common_voice}.

\noindent \textbf{Multi-Task Learning \& Invariances:}
Most highly related to this work are others which deal with multi-task learning and/or the study of invariances/equivariances learnt by self-supervision. In particular, our work relates to: \cite{linus_invariance}, which showed that different computer vision tasks benefit from different (in)variances; HyperSimCLR \cite{hyper_simclr}, which demonstrated that a hypernetwork can adapt a representation to the (in)variances needed for downstream tasks; and AugSelf \cite{aug_self} that also investigates co-learning contrastive and predictive self-supervision in computer vision. Our work differentiates itself in a few key ways, including: a parameter-efficient solution to multi-task learning via the use of adapters \cite{residual_adapters}; the application to acoustic data; the extent and complexity of applied augmentations; and the diversity of downstream tasks considered. Other related works include those which investigate multi-task learning in the audio domain, such as PASE \cite{pase}.

\section{MT-SLVR}
Motivated by the intuition that solely learning invariances to augmentations may be suboptimal for specific downstream tasks, we propose to co-learn opposing objectives. Specifically, we learn a feature space using both contrastive and predictive self-supervision. We name our approach \textbf{MT-SLVR} (\textbf{M}ulti-\textbf{T}ask \textbf{S}elf-Supervised \textbf{L}earning for Transformation In/(\textbf{V}ariant) \textbf{R}epresentations). 
We conjecture that \emph{different downstream tasks benefit from different type and strength of invariance, and that providing both augmentation sensitive and invariant features will lead to superior performance}.

\noindent\textbf{Objective:} We introduce the notation $t_{\phi}(t_{\phi}^{aug})_{aug \in \mathcal{A}}$ to denote applied augmentation pipelines, where $t_{\phi}$ is a composition of individual augmentations ($t_{\phi}^{aug}$) and their parametrisations ($\phi$), and $\mathcal{A}$ is the set of augmentations used during training (e.g. $\mathcal{A} = \{\text{Pitch Shift, Fade}\}$). For our contrastive component ($\mathcal{L}_{Cont}$), we calculate loss in the same manner as the original works \cite{simclr,simsiam}. For the predictive component, we propose a Multi-Label Augmentation Prediction (MLAP) framework, where augmentations are independently predicted for input samples. Formally, given a base feature extractor $f_{\theta}$, a multi-layer MLP for transformation prediction $\psi_{\theta}$, the Binary Cross-Entropy loss (BCE), and augmented samples $v_1 = t_{\phi_{1}}(x)$ and $v_2 = t_{\phi_{2}}(x)$, our predictive loss is defined as:
\begin{equation}
    \mathcal{L}_{MLAP}(x) = \sum_{aug \in \mathcal{A}} BCE\left( \psi_{\theta}(f_{\theta}(v_1), f_{\theta}(v_2)), y_{aug} \right)
\end{equation}
where 
\begin{equation}
    y_{aug} = \mathbb{I}(t_{\phi}^{aug}(x))
\end{equation}
and $\mathbb{I}$ is the indicator function which takes a value of 0 if $t_{\phi}^{aug}$ has not been applied to $x$, and 1 if it has. For a given $x$, sampled augmentation pipelines $v_1$ and $v_2$ consist of the same type and ordering of augmentations, however do not share augmentation specific parameters. This is done to keep alignment with original SimCLR \cite{simclr} and SimSiam \cite{simsiam} works, which also make this restriction. The total objective for the multi-task problem can be expressed as:
\begin{equation}
    \mathcal{L}_{Total} = \mathcal{L}_{Cont} + \lambda \cdot \mathcal{L}_{MLAP}
\end{equation}
Where $\lambda$ is a hyperparameter which balances the individual losses. Optimising for this total objective encourages the shared extractor $f_{\theta}$ to learn both augmentation-invariant and augmentation-sensitive features.

\noindent\textbf{Architecture:} We propose jointly optimising the objectives by utilising both task-specific and task-agnostic features within the neural network. More formally, we introduce the notation $\theta_s$, $\theta_0$ and $\theta_1$ to represent shared, contrastive specific and predictive specific parameters respectively. Objectives for our multi-task approach are then:
\begin{equation}
    \mathcal{L}_{Cont} (x;  f_{\theta_s}, f_{\theta_0}  )
\end{equation}
\begin{equation}
    \mathcal{L}_{MLAP} (x;  f_{\theta_s}, f_{\theta_1}  )
\end{equation}
where the task-specific parameters are defined by architectural changes made to assist multi-task. In particular, we employ two of these changes: 1) Splitting the final output layer of the network such that each task corresponds to the outputs of half of the final layer neurons; and 2) The fitting of residual or batch-normalisation adapters throughout the model, as in \cite{residual_adapters}. We use adapters in the same way as proposed in the original work, where lightweight modules are added around residual blocks. These modules take the form:
\begin{equation}
    g(x ; \alpha)=x+\alpha * x
\end{equation}
where $\alpha$ can either be a batch normalisation or 1x1 convolutional layer. Although lightweight, the included adapters do influence parametrisation. As a multiplier compared to the base model, models fit with adapters have the following parametrisation: Batch Normalisation (BN) $\approx 1\times$, Series Adapters (Series) $1.2 \times$, Parallel Adapters (Parallel) $1.2\times$.

    \setlength{\tabcolsep}{10pt} 
    \renewcommand{\arraystretch}{1} 
    \begin{table}[h]
    
        \caption{Details of augmentations used, along with their respective parameters. We introduce shorthand for later use.}
        
        \centering
        \resizebox{\columnwidth}{!}{%
            \begin{tabular}{cccc}
			 \toprule    
             Augmentaton & Shorthand & Parameter & Value(s)  \\
            \toprule
            Pitch Shift & PS & Min / Max Transpose Semitones & -15 / 15 \\
            
            \midrule
            
            & & Shape & Lin, Log, Exp\\
            \multirow{-2}{*}{Fade} & \multirow{-2}{*}{FD}& Max In / Out Ratio & 0.5 / 0.5 \\
            
            \midrule
            
            & & Min / Max SNR in dB & 3 / 30 \\
            \multirow{-2}{*}{White Noise} & \multirow{-2}{*}{WN} & Min Max f-Decay & -1 / 0  \\
            
            \midrule
            
            & & Min / Max SNR in dB & 3 / 30 \\
            \multirow{-2}{*}{Mixed Noise} & \multirow{-2}{*}{MN} & Min Max f-Decay & -2 / 2 \\

            \midrule

            Time Masking & TM & Max Mask Ratio & 0.125 \\

            \midrule

            Time Shift & TS$^1$ & Min / Max Shift Ratio & 0.5 \\

            \midrule

            Time Stretch & TS$^2$ & Min / Max Stretch Factor & 0.5 / 1.5 \\

             \bottomrule

		    \end{tabular}}
        \label{table:augs}
    \end{table}

%


\noindent\textbf{Augmentations:} We use the augmentations and corresponding parameters from CLAR \cite{clar}, see Table \ref{table:augs}. These encompass seven temporal or frequency-based augmentations. For sampling, we place no restrictions on the number of augmentations per sample, nor in which order they appear, except that at least one augmentation must be present. Each augmentation (except for the first) is activated with its own Bernoulli probability, allowing cases in which all augmentations are present.  


\setlength{\tabcolsep}{20pt} 
\renewcommand{\arraystretch}{0.8} 
\begin{table*}[t]
  \caption{High level details of all datasets considered. Split into environmental sounds (TOP) and different types of speech (BOTTOM). Included datasets originating from MetaAudio are marked with *.}
  \centering
   \resizebox{2\columnwidth}{!}{%
  \begin{tabular}{cccccc}
    \toprule
    Name & Setting & $N^o$ Classes & $N^o$ Samples & Format & Sample Length \\
    \midrule
    Balanced AudioSet \cite{audioset} & Mixed & 527 & 20,550 & Fixed & 10s \\
    \midrule
    ESC-50  \cite{esc} *       & Environmental         & 50   & 2,000   & Fixed    & 5s\\
    NSynth \cite{nsynth} *     & Instrumentation       & 1,006 & 305,978 & Fixed    & 4s \\
    FDSKaggle18 \cite{kaggle_18} *  & Mixed                 & 41   & 11,073  & Variable & 0.3s - 30s \\
    Watkins Marine Mammal Sounds \cite{watkins} *      & Marine Mammals         & 32   & 1,698   & Variable    & 0.1 - 150s\\
    BirdCLEF 2020 (Pruned) \cite{birdclef_2020} * & Bird Song    & 715  & 63,364  & Variable & 3s - 180s \\
    \midrule

    VoxCeleb1  \cite{vox_celeb_1} *  & Speaker              & 1,251 & 153,516 & Variable & 3s - 180s\\
    SpeechCommandsV2 \cite{speech_commands} *   & Keyword       & 35 & 105,829 & Fixed    & 1s\\
    Crema-D \cite{cremad} & Emotion & 6 & 7,442 & Variable & 1s - 5s\\
    Speech Accent Archive \cite{saa} & Accent & 122 & 2,060 & Variable & 17s - 110s \\
    Common Voice v12 Delta \cite{common_voice} & Language & 88 & 256,243 & Variable & 5s - 30s\\
    
    \bottomrule
  \end{tabular}}
  \label{table:datasets}
\end{table*}

\section{Setup}
\label{section:setup}
\noindent\textbf{Pre-Training}: Our pre-training pipeline consists of two distinct parts, self-supervised learning on the balanced training subset of the popular AudioSet \cite{audioset} (containing $\approx 60 hrs$ of audio), and hyperparameter optimisation based on average performance over the validation splits of the MetaAudio benchmark \cite{metaaudio}. More specifically, we selected learning rates for each approach by comparing the average rank of trained models on tasks drawn from MetaAudio. Learning rates tested were between 1x10$^{-6}$ and 1x10$^{-2}$. Rates selected were 1x10$^{-4}$ for baselines and 0.5x10$^{-4}$ for multi-task approaches.  All included models were trained for 1,000 epochs on the ResNet-18 backbone (with a final dense output of 1,000), using the Adam \cite{adam_opt} optimiser. We generate sample-wise augmentations, where 1 to 7 augmentations (see Table \ref{table:augs}) are selected and applied in a random order. Models were trained on a mix of RTX GPUs and on average took 30 hrs to complete. 
\newline\noindent\textbf{Data Processing:} Like other works \cite{clar,cola} we utilise a 2-d 3-channel spectrogram-based representation for input to the model. For pre-training, augmentations are applied before this conversion. For variable length sets at evaluation time, we utilise fixed length splitting and majority voting for classification, as described in \cite{metaaudio}. 
\newline\noindent\textbf{Few-Shot Classification:} We evaluate our models on few-shot classification tasks drawn from a variety of datasets. Within our selection, we consider both the general audio and speech domains. For general audio, we make use of the MetaAudio \cite{metaaudio} benchmark, while for speech we source additional datasets \cite{cremad, saa, common_voice}. For those included in MetaAudio, we use the test split presented by the original work, while for our own speech datasets, we utilise all classes for testing. We detail all of these datasets in Table~\ref{table:datasets}. Following the methodology from \cite{linear_eval}, we freeze our learnt ResNet-18 backbone after pre-training (hence no fine-tuning) and solve tasks using a per few-shot task linear classifier. More specifically, we use a log-loss instantiation of the SGDClassifier as provided in sklearn \cite{sklearn}. For models which have multiple heads, we concatenate features before input to the classifier. Performance on each downstream dataset is reported as the average 5-way 1-shot task performance, $\pm$ the 95\% Confidence Interval (CI), taken over 10,000 tasks. 
\newline\noindent\textbf{Competitors:} We compare the following methods: \underline{Contrastive} learning only \cite{simclr,simsiam}; Multi-label transformation \underline{Predictive} learning only; \underline{MT-Simple} denoting our multi-task loss on a simple ResNet backbone; \underline{MT-Split} denoting a ResNet backbone split at the final layer with one loss applied to each branch; \underline{MT-\{BN, Series, Parallel\}} denoting a parameter-efficient multi-task split with shared ResNet blocks and task-specific BN, Series, or Parallel adapters. We note that we exclude Wav2Vec \cite{wav2vec2} and other Contrastive Predictive Coding (CPC) based methods from our comparison as they do not explicitly learn either augmentation invariances or variances, and hence fall out of scope of our research question.
\newline\noindent\textbf{Invariance Analysis:} We also analyse our model in terms of measuring the learned augmentation (in)variance of the multi-task learned representation.  We follow the work by Ericsson et al. \cite{linus_invariance} by utilising the Mahalanobis distance between our original training samples and their transformed counterparts. Like in \cite{linus_invariance}, given a feature extractor $f$ with feature space covariance $\Sigma$, a transformation $t_{\phi}^{aug}$ whose parameters belong to a set of all possible $\phi \in \Phi$, and a dataset $\mathcal{D}$, we measure strength of invariance as:

\begin{equation}
    M_f^{T_{\Phi}^{aug}}(\mathcal{D})=\frac{1}{|\mathcal{D}||\Phi|} \sum_{x \in \mathcal{D}} \sum_{\phi \in \Phi} m_f^{t_{\phi}^{aug}}(x)
\end{equation}
where 
\begin{equation}
    m_f^{t_{\phi}^{aug}}(x)=\sqrt{\left(f(x)-f\left(v\right)\right) \boldsymbol{\Sigma}^{-1}\left(f(x)-f\left(v\right)\right)^T}
\end{equation}
and $v$ is the transformed input sample $t_{\phi}^{aug}(x)$. A feature extractor with zero total Mahalanobis distance between the original input samples and their transformed counterparts is perfectly invariant, while values greater represent increasing sensitivity.

\section{Results}
\subsection{Few-Shot Learning Results}

    \setlength{\tabcolsep}{10pt} 
    \renewcommand{\arraystretch}{1} 
    \begin{table*}[t]
    
        \caption{5-Way 1-Shot Performance Comparison between \textbf{SimCLR} methods. We compare SimCLR on its own (Baseline), Multi-Task Learning with no, or simple tricks (MT-Simple / Split), and Multi-Task with adapters (MT-Bn / Series / Parallel).}
        
        \centering
        \resizebox{\textwidth}{!}{%
            \begin{tabular}{c|cccccccccc|c}
   
			 \toprule
             Model ($f_\theta$) & ESC-50 & NSynth & Kaggle18 &  Watkins & BirdClef & VoxCeleb & SCv2 & Crema-D & SAA & C-Voice & Avg Rank \\

             \toprule 

            Cont Only & 63.40$^{\pm0.39}$ & 66.44$^{\pm0.40}$ & 37.64$^{\pm0.40}$ & 52.91$^{\pm0.41}$ & \textbf{30.93$^{\pm0.38}$} & 31.18$^{\pm0.37}$ & \textbf{25.68$^{\pm0.35}$} & 29.10$^{\pm0.36}$ & 26.16$^{\pm0.34}$ & 33.33$^{\pm0.38}$ & 3.9\\

            Pred Only & 37.76$^{\pm0.34}$ & 62.52$^{\pm0.36}$ & 21.72$^{\pm0.34}$ & 28.88$^{\pm0.39}$ & 21.04$^{\pm0.35}$ & 21.68$^{\pm0.40}$ & 20.08$^{\pm0.37}$ & 21.68$^{\pm0.33}$ & 23.08$^{\pm0.34}$ & 23.00$^{\pm0.42}$ & 7.0\\

            \midrule

            MT-Simple & 64.23$^{\pm0.39}$ & 66.73$^{\pm0.39}$ & 36.70$^{\pm0.40}$ & 55.26$^{\pm0.42}$ & 29.39$^{\pm0.37}$ & 30.91$^{\pm0.36}$ & 24.02$^{\pm0.34}$ & 29.07$^{\pm0.37}$ & 26.32$^{\pm0.34}$ & 33.21$^{\pm0.38}$ & 4.4\\

            MT-Split & 61.23$^{\pm0.39}$ & 65.29$^{\pm0.40}$ & 33.42$^{\pm0.38}$ & 53.19$^{\pm0.41}$ & 27.38$^{\pm0.36}$ & 29.71$^{\pm0.36}$ & 23.40$^{\pm0.34}$ & 28.66$^{\pm0.37}$ & 26.27$^{\pm0.34}$ & 31.80$^{\pm0.37}$ & 5.8 \\

            MT-Bn & 69.17$^{\pm0.38}$ & \textbf{72.44$^{\pm0.39}$} & \textbf{39.11$^{\pm0.41}$} & 58.80$^{\pm0.43}$ & 30.32$^{\pm0.38}$ & 32.10$^{\pm0.38}$ & 24.40$^{\pm0.35}$ & \textbf{30.03$^{\pm0.38}$} & 28.61$^{\pm0.37}$ & 34.72$^{\pm0.40}$ & 2.1 \\

            MT-Series & 69.00$^{\pm0.39}$ & 71.25$^{\pm0.39}$ & 37.28$^{\pm0.40}$ & 58.92$^{\pm0.42}$ & 28.82$^{\pm0.38}$ & 33.26$^{\pm0.38}$ & 24.66$^{\pm0.35}$ & 29.57$^{\pm0.38}$ & 28.74$^{\pm0.37}$ & 34.23$^{\pm0.38}$ & 2.9\\

            MT-Parallel & \textbf{69.53$^{\pm0.39}$} & 71.81$^{\pm0.39}$ & 38.36$^{\pm0.40}$ & \textbf{59.49$^{\pm0.42}$} & 29.49$^{\pm0.38}$ & \textbf{33.58$^{\pm0.39}$} & 23.65$^{\pm0.34}$ & 29.61$^{\pm0.38}$ & \textbf{28.92$^{\pm0.37}$} & \textbf{35.22$^{\pm0.40}$} & \underline{\textbf{1.9}}\\

             \bottomrule

		    \end{tabular}}
        \label{table:simclr_results}
        \vspace{-5pt}
    \end{table*}

    \setlength{\tabcolsep}{10pt} 
    \renewcommand{\arraystretch}{1} 
    \begin{table*}[t]
    
        \caption{5-Way 1-Shot Performance Comparison between \textbf{SimSiam} methods. We compare SimSiam on its own (Baseline), Multi-Task Learning with no, or simple tricks (MT-Simple / Split), and Multi-Task with adapters (MT-Bn / Series / Parallel).}
        
        \centering
        \resizebox{\textwidth}{!}{%
            \begin{tabular}{c|cccccccccc|c}
   
			 \toprule
             Model ($f_\theta$) & ESC-50 & NSynth & Kaggle18 &  Watkins & BirdClef & VoxCeleb & SCv2 & Crema-D & SAA & C-Voice & Avg Rank \\

             \toprule 

            Cont Only & 51.74$^{\pm0.40}$ & 68.78$^{\pm0.39}$ & 31.72$^{\pm0.37}$ & 48.29$^{\pm0.42}$ & 23.94$^{\pm0.33}$ & 24.13$^{\pm0.32}$ & \textbf{23.80$^{\pm0.35}$} & 28.11$^{\pm0.32}$ & 23.51$^{\pm0.31}$ & 28.50$^{\pm0.36}$ & 5.0\\

           Pred Only & 37.76$^{\pm0.34}$ & 62.52$^{\pm0.36}$ & 21.72$^{\pm0.34}$ & 28.88$^{\pm0.39}$ & 21.04$^{\pm0.35}$ & 21.68$^{\pm0.40}$ & 20.08$^{\pm0.37}$ & 21.68$^{\pm0.33}$ & 23.08$^{\pm0.34}$ & 23.00$^{\pm0.42}$ & 7.0\\

            \midrule

            MT-Simple & 51.87$^{\pm0.40}$ & 69.68$^{\pm0.38}$ & 29.45$^{\pm0.36}$ & 53.13$^{\pm0.42}$ & 24.14$^{\pm0.34}$ & 25.84$^{\pm0.35}$ & 21.81$^{\pm0.33}$ & 26.42$^{\pm0.36}$ & 27.65$^{\pm0.35}$ & 28.96$^{\pm0.36}$ & 4.4\\

            MT-Split &52.07$^{\pm0.40}$ & 68.26$^{\pm0.39}$ & 28.68$^{\pm0.36}$ & 52.04$^{\pm0.42}$ & 24.47$^{\pm0.34}$ & 25.58$^{\pm0.34}$ & 22.08$^{\pm0.33}$ & 26.69$^{\pm0.36}$ & 26.70$^{\pm0.35}$ & 28.58$^{\pm0.36}$ & 4.8\\

            MT-Bn & 58.41$^{\pm0.41}$ & 73.42$^{\pm0.38}$ & 31.69$^{\pm0.39}$ & 55.46$^{\pm0.43}$ & 25.44$^{\pm0.35}$ & 26.71$^{\pm0.36}$ & 21.99$^{\pm0.34}$ & 28.90$^{\pm0.37}$ & 27.38$^{\pm0.35}$ & 29.64$^{\pm0.37}$ & 3.0\\

            MT-Series & 57.24$^{\pm0.40}$ & 74.37$^{\pm0.37}$ & 37.31$^{\pm0.39}$ & 54.70$^{\pm0.42}$ & 25.20$^{\pm0.36}$ & 26.87$^{\pm0.36}$ & 22.64$^{\pm0.34}$ & 30.62$^{\pm0.37}$ & 26.44$^{\pm0.35}$ & 31.07$^{\pm0.38}$ & 2.7 \\

            MT-Parallel & \textbf{60.61$^{\pm0.41}$} & \textbf{76.36$^{\pm0.37}$} & \textbf{37.59$^{\pm0.40}$} & \textbf{57.98$^{\pm0.42}$} & \textbf{25.45$^{\pm0.37}$} & \textbf{28.66$^{\pm0.37}$} & 23.08$^{\pm0.34}$ & \textbf{30.72$^{\pm0.37}$} & \textbf{27.94$^{\pm0.36}$} & \textbf{32.72$^{\pm0.38}$} & \underline{\textbf{1.1}}\\
            
             \bottomrule


		    \end{tabular}}
        \label{table:simsiam_results}
        \vspace{-5pt}
    \end{table*}

Across experiments (see Tables \ref{table:simclr_results} and \ref{table:simsiam_results}), we observe strong improvements over both baselines (contrastive and predictive only), across all datasets. Ranked, the top 3 consist of the batch-normalisation, series and parallel adapters, followed by a mix of the others. Notably, both the naive multi-task approach (MT-Simple), where all features are shared between tasks, and the split branch counterpart (MT-Split) both yield worse (SimCLR) or only marginal improvements (SimSiam) on the baseline contrastive approaches. This shows that a richer multi-task architecture is necessary, and our parallel adapter approach provides this. We also observe some differences between contrastive methods used. Specifically, for SimCLR we observe a much higher spread of top ranking methods, while for SimSiam the parallel adapter method performs best in 9/10 cases, typically with much larger margins between it and the next best. We also observe that out of all 10 datasets, our absolute top performances in 8/10 are from SimCLR based methods.

    \setlength{\tabcolsep}{10pt} 
    \renewcommand{\arraystretch}{1} 
    \begin{table}[h]
        \centering
        \caption{Measured average Mahalanobis distance between original and augmented training (AudioSet) samples for \textbf{SimCLR} based models' (P)redictive or (C)ontrastive heads. Lower values indicate more invariance to the transformation. }

        \resizebox{\columnwidth}{!}{%
            \begin{tabular}{c|cccccccc|c}
   
			 \toprule
             Model ($f_\theta$) & Head & PS & FD & WN &  MN & TM & TS$^1$ & $TS^2$ & Avg\\

             \toprule 

            Cont Only & - & 32.01 & 31.23 & 31.67 & 31.66 & 30.96 & 31.61 & 31.35 & 31.5 \\

            \midrule
            
            Pred Only & - & 38.17 & 40.81 & 41.26 & 40.73 & 54.21 & 30.39 & 43.83 & 41.34\\

            \midrule

            MT-Simple & - & 29.09 & 32.62 & 31.86 & 31.01 & 29.05 & 22.45 & 36.77 & 30.41 \\

            \midrule
            
            & C & 37.22 & 38.66 & 39.16 & 37.86 & 37.50 & 30.77 & 43.01 & 37.74 \\
            \multirow{-2}{*}{MT-Split} & P & 37.31 & 38.73 & 39.24 & 37.95 & 37.78 & 30.77 & 43.15 & 37.85 \\

            \midrule
            
            & C & 27.62 & 31.93 & 28.48 & 27.51 & 29.98 & 21.38 & 35.68 & 28.94 \\
            \multirow{-2}{*}{MT-Bn} & P & 29.58 & 34.95 & 31.98 & 30.79 & 35.39 & 21.42 & 40.67 & 32.11 \\

            \midrule
            
            & C &22.71 & 26.16 & 23.38 & 23.15 & 21.00 & 21.33 & 30.24 & 24.00 \\
            \multirow{-2}{*}{MT-Series} & P & 36.59 & 37.71 & 39.09 & 38.78 & 44.44 & 21.28 & 42.12 & 37.14 \\
            
            \midrule
            
            & C & 31.41 & 32.37 & 31.69 & 31.51 & 30.07 & 30.34 & 33.88 & 31.61 \\
            \multirow{-2}{*}{MT-Parallel} & P & 35.67 & 42.53 & 39.99 & 40.01 & 42.74 & 30.24 & 41.21 & 38.91 \\

             \bottomrule

		    \end{tabular}}
        \label{table:simclr_inv}
        \vspace{-6pt}
    \end{table}
    \setlength{\tabcolsep}{10pt} 
    \renewcommand{\arraystretch}{1.1} 
    \begin{table}[h]
    
        \caption{Average linear classifier feature weight for the (P)redictive and (C)ontrastive heads in multi-task 
        \textbf{SimCLR}.}
        
        \centering
        \resizebox{\columnwidth}{!}{%
            \begin{tabular}{c|cccccccc|c}
   
			 \toprule
             Model ($f_\theta$) & Head & ESC-50 & NSynth & BirdClef &  Crema-D & SAA & C-Voice\\

             \toprule 

            & C & 0.43& 0.41 &0.40 &0.46& 0.41 &0.36 \\
            \multirow{-2}{*}{MT-Split} & P & 0.57& 0.59 &0.60 &0.54& 0.59 &0.64 \\

            \midrule
            
            & C & 0.41& 0.39 &0.41 &0.43& 0.40 &0.37 \\
            \multirow{-2}{*}{MT-Bn} & P & 0.59& 0.61 &0.59 &0.57& 0.60 &0.63 \\

            \midrule
            
            & C &0.39& 0.33 &0.36 &0.39& 0.37 &0.30 \\
            \multirow{-2}{*}{MT-Series} & P & 0.61& 0.67 &0.64 &0.61& 0.63 &0.70 \\
            
            \midrule
            
            & C & 0.41& 0.37 &0.36 &0.38& 0.38 &0.31 \\
            \multirow{-2}{*}{MT-Parallel} & P & 0.59& 0.63 &0.64 &0.62& 0.62 &0.69 \\

             \bottomrule

		    \end{tabular}}
        \label{table:simclr_weights}
        \vspace{-13pt}
    \end{table}

\subsection{Invariance Analysis}
To illustrate what (in)variances our framework has learned, we measure the distance between  original and augmented samples (Sec \ref{section:setup}) for our training set. The results in Tab.~\ref{table:simclr_inv} show a few key trends. In particular, we note that: 1) Different heads of our multi-task approaches do indeed learn significantly different degrees of invariance to applied augmentations; and 2) On average, even the simple multi-task approaches decrease invariance strength compared to the contrastive baseline. Interestingly, we observe that the naive multi-task baselines (MT-Simple, MT-Split) do not successfully learn distinct invariances in either case, which may explain their weaker performance relative to other proposed approaches. We do not see a clear trend where a larger difference in augmentation strength between heads is predictive of final performance ranking. For example, the series adapter has the largest invariance strength difference, however does not rank first for either contrastive framework. Thus, although diverse (in)variance strength is important in providing a flexible representation, there is a more complex relationship that still needs to be understood.  Finally, we expand our analysis by considering the average weight norms learned for each of the multi-task heads by our linear classifier for a representative set of datasets in Tab.~\ref{table:simclr_weights}.Our results illustrate that across different downstream tasks, the relative importance of contrastive versus predictive heads varies. This illustrates why the presence of both is advantageous for the numerical results in Tab~\ref{table:simclr_results}, and shows how downstream tasks can easily tune the degree of importance attributed to each feature by learning the linear combination, removing the need for human intervention at either the pre-train or downstream task steps.

\section{Conclusion \& Future Work}
We considered the idea that different downstream tasks may prefer different degrees of (in)variance in a pre-trained representation. Leveraging this insight, we developed a novel multi-task learner that exploits both contrastive and predictive learning, providing both augmentation invariant and augmentation sensitive features. To this end, we developed a novel multi-task architecture that provides both features by sharing most parameters and exploiting compact task-specific adapters. Our analysis showed that this multi-task architecture indeed learns substantially different invariances with each head. Each downstream task learning a linear combination of these features, is free to select its own operating point on the (in)variance spectrum., reducing the need for specific pre-train to downstream task tuning. We evaluated our approach on a diverse suite of few-shot classification tasks from a total of 10 audio and speech datasets and two contrastive learners (SimSiam and SimCLR). The results showed that our multi-task features improve on pure contrastive learning and provides the best performance in nearly all cases. In particular, we highlight that SimCLR with parallel adapters performed best on average. This work showed that multi-task learning produces more general features. This will enable faster adaptation to diverse downstream applications where lots of labelled data is not available, such as for voice recognition, speaker identification and emotion detection.

\section{Acknoledgement}
This work is supported by the Engineering and Physical
Sciences Research Council of the UK (EPSRC) Grant number
EP/S000631/1 and the UK MOD University Defence Research
Collaboration (UDRC) in Signal Processing, EPSRC iCASE account EP/V519674/1 and Thales UK Ltd. 

\newpage
\bibliographystyle{IEEEtran}
\bibliography{mybib}

\begin{thebibliography}{10}
\providecommand{\url}[1]{#1}
\csname url@samestyle\endcsname
\providecommand{\newblock}{\relax}
\providecommand{\bibinfo}[2]{#2}
\providecommand{\BIBentrySTDinterwordspacing}{\spaceskip=0pt\relax}
\providecommand{\BIBentryALTinterwordstretchfactor}{4}
\providecommand{\BIBentryALTinterwordspacing}{\spaceskip=\fontdimen2\font plus
\BIBentryALTinterwordstretchfactor\fontdimen3\font minus
  \fontdimen4\font\relax}
\providecommand{\BIBforeignlanguage}[2]{{%
\expandafter\ifx\csname l@#1\endcsname\relax
\typeout{** WARNING: IEEEtran.bst: No hyphenation pattern has been}%
\typeout{** loaded for the language `#1'. Using the pattern for}%
\typeout{** the default language instead.}%
\else
\language=\csname l@#1\endcsname
\fi
#2}}
\providecommand{\BIBdecl}{\relax}
\BIBdecl

\bibitem{simclr}
T.~Chen, S.~Kornblith, M.~Norouzi, and G.~Hinton, ``A simple framework for
  contrastive learning of visual representations,'' in \emph{International
  conference on machine learning}.\hskip 1em plus 0.5em minus 0.4em\relax PMLR,
  2020, pp. 1597--1607.

\bibitem{linus_invariance}
L.~Ericsson, H.~Gouk, and T.~Hospedales, ``Why do self-supervised models
  transfer? on the impact of invariance on downstream tasks,'' in \emph{33rd
  British Machine Vision Conference 2022, {BMVC} 2022, London, UK, November
  21-24, 2022}.\hskip 1em plus 0.5em minus 0.4em\relax {BMVA} Press, 2022.

\bibitem{aug_self}
H.~Lee, K.~Lee, K.~Lee, H.~Lee, and J.~Shin, ``Improving transferability of
  representations via augmentation-aware self-supervision,'' \emph{Advances in
  Neural Information Processing Systems}, vol.~34, pp. 17\,710--17\,722, 2021.

\bibitem{hyper_simclr}
R.~Chavhan, J.~Stuehmer, C.~Heggan, M.~Yaghoobi, and T.~Hospedales, ``Amortised
  invariance learning for contrastive self-supervision,'' in
  \emph{International Conference on Learning Representations}, 2023.

\bibitem{clar}
H.~Al-Tahan and Y.~Mohsenzadeh, ``Clar: Contrastive learning of auditory
  representations,'' in \emph{International Conference on Artificial
  Intelligence and Statistics}.\hskip 1em plus 0.5em minus 0.4em\relax PMLR,
  2021, pp. 2530--2538.

\bibitem{linus_survey}
L.~Ericsson, H.~Gouk, C.~C. Loy, and T.~M. Hospedales, ``Self-supervised
  representation learning: Introduction, advances, and challenges,'' \emph{IEEE
  Signal Processing Magazine}, vol.~39, no.~3, pp. 42--62, 2022.

\bibitem{ssl_audio_survey}
S.~Liu, A.~Mallol-Ragolta, E.~Parada-Cabaleiro, K.~Qian, X.~Jing, A.~Kathan,
  B.~Hu, and B.~W. Schuller, ``Audio self-supervised learning: A survey,''
  \emph{Patterns}, vol.~3, no.~12, p. 100616, 2022.

\bibitem{ssl_speech_survey}
A.~rahman Mohamed, H.~yi~Lee, L.~Borgholt, J.~D. Havtorn, J.~Edin, C.~Igel,
  K.~Kirchhoff, S.-W. Li, K.~Livescu, L.~Maal{\o}e, T.~N. Sainath, and
  S.~Watanabe, ``Self-supervised speech representation learning: A review,''
  \emph{IEEE Journal of Selected Topics in Signal Processing}, vol.~16, pp.
  1179--1210, 2022.

\bibitem{simsiam}
X.~Chen and K.~He, ``Exploring simple siamese representation learning,'' in
  \emph{Proceedings of the IEEE/CVF Conference on Computer Vision and Pattern
  Recognition}, 2021, pp. 15\,750--15\,758.

\bibitem{moco}
K.~He, H.~Fan, Y.~Wu, S.~Xie, and R.~Girshick, ``Momentum contrast for
  unsupervised visual representation learning,'' in \emph{2020 IEEE/CVF
  Conference on Computer Vision and Pattern Recognition (CVPR)}, 2020, pp.
  9726--9735.

\bibitem{barlow_twins}
J.~Zbontar, L.~Jing, I.~Misra, Y.~LeCun, and S.~Deny, ``Barlow twins:
  Self-supervised learning via redundancy reduction,'' in \emph{Proceedings of
  the 38th International Conference on Machine Learning, {ICML} 2021}, ser.
  Proceedings of Machine Learning Research, M.~Meila and T.~Zhang, Eds., vol.
  139.\hskip 1em plus 0.5em minus 0.4em\relax {PMLR}, 2021, pp.
  12\,310--12\,320.

\bibitem{rot_net}
S.~Gidaris, P.~Singh, and N.~Komodakis, ``Unsupervised representation learning
  by predicting image rotations,'' in \emph{International Conference on
  Learning Representations}, 2018.

\bibitem{cola}
A.~Saeed, D.~Grangier, and N.~Zeghidour, ``Contrastive learning of
  general-purpose audio representations,'' in \emph{ICASSP 2021-2021 IEEE
  International Conference on Acoustics, Speech and Signal Processing
  (ICASSP)}.\hskip 1em plus 0.5em minus 0.4em\relax IEEE, 2021, pp. 3875--3879.

\bibitem{ucl_ser}
E.~Fonseca, D.~Ortego, K.~McGuinness, N.~E. O’Connor, and X.~Serra,
  ``Unsupervised contrastive learning of sound event representations,'' in
  \emph{ICASSP 2021-2021 IEEE International Conference on Acoustics, Speech and
  Signal Processing (ICASSP)}.\hskip 1em plus 0.5em minus 0.4em\relax IEEE,
  2021, pp. 371--375.

\bibitem{transient}
S.-Y. Chou, K.-H. Cheng, J.-S.~R. Jang, and Y.-H. Yang, ``Learning to match
  transient sound events using attentional similarity for few-shot sound
  recognition,'' in \emph{ICASSP 2019 - 2019 IEEE International Conference on
  Acoustics, Speech and Signal Processing (ICASSP)}, 2019, pp. 26--30.

\bibitem{fs_fsd}
Y.~Wang, N.~J. Bryan, J.~Salamon, M.~Cartwright, and J.~P. Bello, ``Who calls
  the shots? rethinking few-shot learning for audio,'' in \emph{2021 IEEE
  Workshop on Applications of Signal Processing to Audio and Acoustics
  (WASPAA)}.\hskip 1em plus 0.5em minus 0.4em\relax IEEE, 2021, pp. 36--40.

\bibitem{metaaudio}
C.~Heggan, S.~Budgett, T.~M. Hospedales, and M.~Yaghoobi, ``Metaaudio: A
  few-shot audio classification benchmark,'' in \emph{ICANN}, 2022.

\bibitem{cremad}
H.~Cao, D.~G. Cooper, M.~K. Keutmann, R.~C. Gur, A.~Nenkova, and R.~Verma,
  ``Crema-d: Crowd-sourced emotional multimodal actors dataset,'' \emph{IEEE
  Transactions on Affective Computing}, vol.~5, no.~4, pp. 377--390, 2014.

\bibitem{saa}
S.~Weinberger, ``Speech accent archive,'' 2015.

\bibitem{common_voice}
R.~Ardila, M.~Branson, K.~Davis, M.~Kohler, J.~Meyer, M.~Henretty, R.~Morais,
  L.~Saunders, F.~Tyers, and G.~Weber, ``Common voice: A massively-multilingual
  speech corpus,'' in \emph{Proceedings of the Twelfth Language Resources and
  Evaluation Conference}.\hskip 1em plus 0.5em minus 0.4em\relax European
  Language Resources Association, 2020, pp. 4218--4222.

\bibitem{residual_adapters}
S.-A. Rebuffi, H.~Bilen, and A.~Vedaldi, ``Learning multiple visual domains
  with residual adapters,'' in \emph{Advances in Neural Information Processing
  Systems}, I.~Guyon, U.~V. Luxburg, S.~Bengio, H.~Wallach, R.~Fergus,
  S.~Vishwanathan, and R.~Garnett, Eds., vol.~30.\hskip 1em plus 0.5em minus
  0.4em\relax Curran Associates, Inc., 2017.

\bibitem{pase}
M.~Ravanelli, J.~Zhong, S.~Pascual, P.~Swietojanski, J.~Monteiro, J.~Trmal, and
  Y.~Bengio, ``Multi-task self-supervised learning for robust speech
  recognition,'' in \emph{ICASSP 2020-2020 IEEE International Conference on
  Acoustics, Speech and Signal Processing (ICASSP)}.\hskip 1em plus 0.5em minus
  0.4em\relax IEEE, 2020, pp. 6989--6993.

\bibitem{audioset}
J.~F. Gemmeke, D.~P.~W. Ellis, D.~Freedman, A.~Jansen, W.~Lawrence, R.~C.
  Moore, M.~Plakal, and M.~Ritter, ``Audio set: An ontology and human-labeled
  dataset for audio events,'' in \emph{ICASSP}, 2017.

\bibitem{esc}
K.~J. Piczak, ``{ESC}: {Dataset} for {Environmental Sound Classification},'' in
  \emph{Proceedings of the 23rd {Annual ACM Conference} on {Multimedia}}, 2015.

\bibitem{nsynth}
J.~Engel, A.~R. Cinjon~Resnick, S.~Dieleman, D.~Eck, K.~Simonyan, and
  M.~Norouzi, ``Neural audio synthesis of musical notes with wavenet
  autoencoders,'' 2017.

\bibitem{kaggle_18}
E.~Fonseca, M.~Plakal, F.~Font, D.~P.~W. Ellis, X.~Favory, J.~Pons, and
  X.~Serra, ``General-purpose tagging of freesound audio with audioset labels:
  Task description, dataset, and baseline,'' in \emph{Proceedings of the DCASE
  2018 Workshop (2018)}.

\bibitem{watkins}
L.~Sayigh, M.~Daher, J.~Allen, H.~Gordon, K.~Joyce, C.~Stuhlmann, and P.~Tyack,
  ``The watkins marine mammal sound database: An online, freely accessible
  resource,'' vol.~27, 01 2016, p. 040013.

\bibitem{birdclef_2020}
S.~Kahl, M.~Clapp, W.~Hopping, H.~Goëau, H.~Glotin, R.~Planqué, W.-P.
  Vellinga, and A.~Joly, ``Overview of birdclef 2020: Bird sound recognition in
  complex acoustic environments,'' 2020.

\bibitem{vox_celeb_1}
A.~Nagrani, J.~S. Chung, and A.~Zisserman, ``Voxceleb: a large-scale speaker
  identification dataset,'' in \emph{INTERSPEECH}, 2017.

\bibitem{speech_commands}
P.~Warden, ``Speech commands: A dataset for limited-vocabulary speech
  recognition,'' 2018.

\bibitem{adam_opt}
D.~P. Kingma and J.~Ba, ``Adam: {A} method for stochastic optimization,'' in
  \emph{3rd International Conference on Learning Representations, {ICLR} 2015,
  San Diego, CA, USA, May 7-9, 2015, Conference Track Proceedings}, Y.~Bengio
  and Y.~LeCun, Eds., 2015.

\bibitem{linear_eval}
S.~Kornblith, J.~Shlens, and Q.~V. Le, ``Do better imagenet models transfer
  better?'' in \emph{2019 IEEE/CVF Conference on Computer Vision and Pattern
  Recognition (CVPR)}.\hskip 1em plus 0.5em minus 0.4em\relax IEEE Computer
  Society, 2019, pp. 2656--2666.

\bibitem{sklearn}
L.~Buitinck, G.~Louppe, M.~Blondel, F.~Pedregosa, A.~Mueller, O.~Grisel,
  V.~Niculae, P.~Prettenhofer, A.~Gramfort, J.~Grobler, R.~Layton,
  J.~VanderPlas, A.~Joly, B.~Holt, and G.~Varoquaux, ``{API} design for machine
  learning software: experiences from the scikit-learn project,'' in \emph{ECML
  PKDD Workshop: Languages for Data Mining and Machine Learning}, 2013.

\bibitem{wav2vec2}
A.~Baevski, Y.~Zhou, A.~Mohamed, and M.~Auli, ``wav2vec 2.0: A framework for
  self-supervised learning of speech representations,'' \emph{Advances in
  neural information processing systems}, vol.~33, pp. 12\,449--12\,460, 2020.

\end{thebibliography}
\end{document}